\documentclass[
]{ceurart}

\usepackage{listings}
\usepackage{tcolorbox}
\usepackage{fontawesome5}
\usepackage{float}

\begin{document}

\copyrightyear{2025}
\copyrightclause{Copyright for this paper by its authors.
  Use permitted under Creative Commons License Attribution 4.0
  International (CC BY 4.0).}
\conference{Preprint - This contribution was accepted at JURIX AI4A2J Workshop 2025}

\title{CourtPressGER: A German Court Decision to Press Release Summarization Dataset}

\author[1]{Sebastian Nagl}
\cormark[1]
\address[1]{Technical University of Munich (TUM)}

\author[1]{Mohamed Elganayni}

\author[1]{Melanie Pospisil}

\author[1]{Matthias Grabmair}

\cortext[1]{Corresponding author: sebastian.nagl@tum.de. The dataset and evaluation framework is available \href{https://github.com/SebastianNagl/CourtPressGER}{via GitHub}.}

\begin{abstract}
  Official court press releases from Germany's highest courts present and explain judicial rulings to the public, as well as to expert audiences. Prior NLP efforts emphasize technical headnotes, ignoring citizen-oriented communication needs. We introduce CourtPressGER, a 6.4k dataset of triples: rulings, human-drafted press releases, and synthetic prompts for LLMs to generate comparable releases. This benchmark trains and evaluates LLMs in generating accurate, readable summaries from long judicial texts. We benchmark small and large LLMs using reference-based metrics, factual-consistency checks, LLM-as-judge, and expert ranking. Large LLMs produce high-quality drafts with minimal hierarchical performance loss; smaller models require hierarchical setups for long judgments. Initial benchmarks show varying model performance, with human-drafted releases ranking highest.
\end{abstract}


\maketitle

\section{Introduction} \label{introduction}

High-level German courts make decisions accessible through press releases that summarize essential aspects and implications in understandable form. Releases authored by judges contain legal authority and lay-friendly narrative, serving as an interface between judiciary and public. This represents targeted legal summarization, where gold data is typically sparse. LLM progress suggests high-quality automatic drafts are within reach, yet robust evaluations of legal decision summaries remain difficult, especially in non-English languages. CourtPressGER advances German legal summarization by:

\begin{enumerate}
\def\labelenumi{\arabic{enumi}.}
\item
  Collecting a large aligned corpus of German decisions and press releases (6.4k pairs),
\item
  deriving decision-specific summarization prompts,
\item
  benchmarking open and commercial LLMs, and
\item
  analyzing performance through automatic and expert assessment
\end{enumerate}

This work contributes a dataset and evaluation framework for German legal summarization research, with initial benchmarks establishing baseline performance.

\section{Related Work}\label{sec:related-work}

Legal‐text summarization has progressed from early sentence-ranking heuristics borrowed from news
\citep{grover2004holj,polsley2016casesummarizer}
to domain-adapted encoder–decoder transformers such as \emph{Legal-BART} and \emph{Legal-PEGASUS} \citep{chalkidis2019deep,zhang2020,aumiller2022eur}. Recent surveys report steady ROUGE gains but emphasize three persistent challenges—extreme document length, jurisdiction-specific jargon, and the absence of factual-consistency metrics \citep{kanapala2019text,akter2025comprehensive}. Researchers address the length issue with hierarchical encoders and chunk-merge strategies for book-length opinions \citep{chang2024} and Indian Supreme Court cases \citep{deroy2024ensemble}. Yet expert evaluations reveal that higher ROUGE scores do not necessarily align with legal usefulness \citep{steffes2023evaluating}, underscoring the need for multi-faceted assessment.

\paragraph{Datasets.}%
Larger legal summarization corpora typically pair expert-targeted summaries with their underlying document. A distinction can be made between: first, summarizing legislation, such as BillSum (U.S.\ bills; \citealp{kornilova2019billsum}) and EUR-Lex-Sum (EU legislation; \citealp{aumiller2022eur}) and, second, case/judgment summarization, such as the US-English Multi-LexSum (U.S.\ civil-rights cases; \citealp{shen2022multi}, including e.g., complaints and motions) and Portuguese BrazilianBR (STF rulings; \citealp{feijo2023improving}). For German, \emph{LegalSum} covers $\sim$100k rulings with legal-holding-focused \emph{Leits\"atze} \citep{glaser2021} from the German legal context, and \citet{rolshoven2024} provides 57k \emph{Regesten} from Swiss courts. Both target legal practitioners and their summaries consist of concise, technical, often extractive headnotes. To date, no corpus of significant size aligns German decisions with non-headnote-based summaries written for mixed audiences, including journalists and the general public.

Outside Germany, few resources focus on citizen-oriented summaries, e.g., TL;DR software license synopses \citep{manor2019plain}, Canadian lay summaries \citep{salaun2022conditional}, and argument-aware rewriting \citep{elaraby2022arglegalsummimprovingabstractivesummarization}.
The  \href{https://www.bmj.de/DE/themen/digitales/digitalisierung_justiz/digitalisierungsinitiative/laendervorhaben/_doc/artikel_vorhaben_14_ALeKS.html\#:~:text=K\%C3\%BCnftig\%20sollen\%20mehr\%20Gerichtsentscheidungen\%20in,Automatisierung\%20vereinfacht}{German ALeKS
project} seeks to automate headnote generation.

Extractive approaches to summarization have been mostly succeeded by abstractive summarization using transformers
\citep{shukla2022legal,moro2022semantic, santosh2025coperlexcontentplanningeventbased, santosh2024lexsumm}
and faithfulness-enhancing rerankers \citep{feijo2023improving}. 
Cross-jurisdiction transfer of smaller models primarily trained in one jurisdiction poses a challenge \citet{santosh2024beyond}. Large commercial models are marketed as capable of summarizing judgments well, which we evaluate through expert analysis.

\paragraph{Evaluation.}%
In addition to classic NLP metrics, we see newer factual metrics like QAGS (Question Answering for evaluating Generated Summaries) (\citet{wang2020}) and FactCC (Factual Consistency Check) (\citet{Kryscinski2019}).
QAGS generates questions from one text and then compares the answers to verify factual correctness. FactCC extracts claims from the one text and checks them against another body. A total factual consistency score is computed from these checks.

Also working on German court press releases, \citet{steffes2023evaluating} demonstrate that ROUGE scores alone fails to capture whether legally salient content is present in a summary. Alternative protocols generate question–answer pairs from the reference or enlist large-language-models as judges, both correlating better with expert panels \citep{xu2021toward}. Current research still lacks (i) German press-release data, (ii) long-context benchmarks in German legal datasets, and
(iii) holistic evaluation beyond ROUGE.

\paragraph{Our contribution.} \textbf{CourtPressGER} addresses these gaps by releasing 6.4k aligned triplets of federal-court decisions, their official press releases, and a synthetic contextual prompt describing the structure of the press release to an LLM. We benchmark six open-source and commercial LLMs using overlap-, embedding- and entailment-based metrics, and validate automatic scores against expert spot-checks. The corpus and baselines establish the first citizen-oriented benchmark for German judicial communication. By supplying a public dataset and a multi-dimensional evaluation suite centered on citizen-oriented summaries, CourtPressGER complements prior resources focused on technical headnotes and narrow evaluation settings, opening a new avenue for research on transparent court communication.

\section{CourtPressGER}\label{courtpressger}

Our dataset includes 6,432 court decisions and corresponding press releases
from Germany's highest courts from the years 1995 to 2023:
Federal Labor Law Court (Bundesarbeitsgericht - BAG), Federal Fiscal Court (Bundesfinanzhof - BFH),
  Federal Court of Justice (Bundesgerichtshof - BGH), Federal Social Court (Bundessozialgericht - BSG),
  Federal Constitutional Court (Bundesverfassungsgericht- BVerfG) and the
  Federal Administrative Court (Bundesverwaltungsgericht - BVerwG).
The dataset and codebase will be publicly available on HuggingFace and GitHub upon acceptance.

For our experiments, we divided the dataset into training, validation, and test splits in an 72.2/11.6/16.3 ratio. The training set contains 4,643 pairs, while the validation set has 744 pairs, and test set has 1,045 pairs. We decided to split chronologically because otherwise the distribution shifts incurred by rotating press office personnel over time would not be captured in the data split, leading to a potential overestimation of performance on unseen data. Descriptive statistics of the cleaned dataset appear in
\hyperref[tab:descriptive_statistics]{Table 1}. 
Using EuroBERT tokenizer, we see decisions averaging 10,810 tokens and
press releases with an average 1,402 tokens with high standard deviations of 10,739 tokens for judgements and 955 tokens for press releases.

\begin{table*}
\centering
\footnotesize             
\setlength{\tabcolsep}{4pt}
\begin{tabular}{@{\extracolsep{\fill}}lrrrrrr}
\toprule
 & \multicolumn{3}{c}{Press Release} & \multicolumn{3}{c}{Judgment} \\
\cmidrule(lr){2-4}\cmidrule(lr){5-7}
Court & Mean & Std & Count & Mean & Std & Count \\
\midrule
BAG     & 1056.37 &  407.50 &  177 & 14148.00 &  7913.64 &  177 \\
BFH          &  800.28 &  213.58 &  761 &  7378.97 &  4410.79 &  761 \\
BGH        & 1386.84 &  680.10 & 2407 &  8216.82 &  5686.26 & 2407 \\
BSG      & 1146.66 &  484.69 &  161 & 11790.02 &  4850.29 &  161 \\
BVerfG & 2039.50 & 1353.63 & 1771 & 14781.53 & 16844.62 & 1771 \\
BVerwG &  942.91 &  336.86 & 1155 & 11734.63 &  8110.92 & 1155 \\
\midrule
\textbf{Overall avg} & 1402.32 &  954.52 &  -- & 10809.58 & 10739.27 &  -- \\
\bottomrule
\end{tabular}
\caption{Statistics of press releases and judgments by court}
\label{tab:descriptive_statistics}
\end{table*}

\section{Experimental Setup}\label{experimental-setup}

\subsection{Synthetic Prompts}\label{synthetic-prompts}
For each decision-press release pair, we generated synthetic prompts through the Anthropic API (Claude Sonnet 3.7) to serve as input for LLMs to generate press releases. These prompts enable LLMs to generate text matching the reference press release in length, intended audience, and topical coverage, making comparison more meaningful than using a single generic prompt across all datapoints from different authors.

We acknowledge that using Claude-generated prompts introduces a potential confounding variable, as prompt quality was not independently benchmarked. We selected Claude after initial testing showed it produced consistently systematic prompts. While model-specific prompt optimization might improve individual results, our uniform approach avoids favoritism and maintains valid relative comparisons between models.

\subsection{Press Release Generation}\label{press-release-generation}
Our pipeline sends the synthetic prompts and the case to the models, collect the generated press releases, and stores them alongside the actual press releases. The models we leverage for this generation are GPT-4o, Llama-3-70B, Teuken-7B\footnote{Our experiments were conducted with version 0.4 of the Teuken7B model. After version 0.6 became available, we updated results for 0-shot and fine-tuned conditions, but redoing the human evaluation was not possible.}, Llama-3-8B, EuroLLM-9B, and Mistral-7B, representing different size classes (7B to 70B parameters) and architectures (both commercial and open-source) available at the time of study. This selection allows us to evaluate performance across the spectrum of model capabilities while maintaining practical feasibility for research deployment.

\subsubsection{Context Limitation}\label{context-limitation}
As expected, context window size limits models' ability to generate high-quality press releases. Models with larger context windows (e.g., GPT-4o) can process entire court decisions at once, while smaller models require document chunking and hierarchical summarization. For decisions exceeding the context window, we implemented a hierarchical summarization approach allowing models to consider the entire document while respecting context limitations.

\subsection{Hierarchical Summarization}\label{hierarchical-summarization}

Since court rulings often exceed context windows, we use hierarchical summarization inspired by \citet{chang2024} to progressively condense text rather than relying on a single pass. While incremental chunk‑by‑chunk updating preserves detail, it can induce coherence errors for large documents \citep{chang2024}. In contrast, hierarchical merging systematically integrates partial synopses, yielding more coherent summaries.

An advantage of our multi‑level procedure is its \emph{abstractive} nature. Following \citet{deroy2024}, we note that generative legal-domain models outperform extractive baselines on metrics such as ROUGE, METEOR, and BERTScore for court cases. It condenses complex legal reasoning into more readable, information‑dense summaries. By generating chunk‑wise abstractive synopses, we want to preserve salient entities and mitigate hallucination risks \citep{deroy2024} through controlled chunk sizes and cross‑chunk aggregation.

Practically, we split each decision into paragraphs, then merge them until reaching a token threshold compatible with the chosen LLM. Each chunk undergoes a “Level‑0” summary to capture key arguments, reasoning, and legal points, after which partial summaries are recursively merged in a structured tree‑like fashion.

\subsection{Instruction-Tuned \textsc{Teuken-7B}: Joint Hierarchical
Summarization and Press-Release Generation}\label{ft-teuken}
Anticipating a negative impact of hierarchical summarization over single-pass summarization, we want to explore how far a smaller model can be improved when specifically trained for this task in a hierarchical fashion, as we have a training set available. We instruction-tune the open-source \textsc{Teuken-7B} model via a \emph{two-stage} approach: first generating faithful level-by-level summaries of long rulings and then rewriting the highest-level summary into a press release.

\paragraph{Stage 1: Hierarchical summarisation.}
For every paragraph chunk \(c_{i}\) of a decision (max.\ token budget \(4\,096\)),  we provide an \emph{explicit instruction} indicating the desired abstraction level. Reference summaries are produced with \texttt{Llama-3-70B-Instruct}, yielding \textbf{35,k} chunk–instruction pairs. We adopt the \emph{stacked} SFT scheme of \citet{pareja2024}, training \textsc{Teuken-7B} on all hierarchy levels simultaneously, which generalizes better than sequential tuning.

\paragraph{Stage 2: Press-release generation.}
We retain each final hierarchical summary \(s^{\text{final}}\) and combine it with the synthetic prompt Section~\ref{synthetic-prompts} and the gold press release \(p^{\star}\). The model is then further trained with a constant learning-rate schedule to map the concatenated input
\((\text{prompt}\!+\!s^{\text{final}})\) to \(p^{\star}\).

\paragraph{Training set-up.}
Unless stated otherwise, we follow \citet{pareja2024}, reducing the effective batch size to 512 and training for four epochs over the mixed corpus, with early stopping on validation perplexity.

\begin{table*}[th]
\centering
\scriptsize
\setlength{\tabcolsep}{2pt}      
\begin{tabular*}{\textwidth}{@{\extracolsep{\fill}}l*{14}{r}}
\toprule
Model &
R1 & B1 & MTR & BERT &
FCC & QAGS &
LJ\_Fact &
LJ\_Compl &
LJ\_Clar &
LJ\_Struc &
LJ\_Ref &
LJ\_Tot &
Human Avg 
\\ \midrule
gpt 4o\_hier & 0.3584 & 0.2275 & 0.1836 & 0.7711 & 0.4915 & 0.2637 & \textbf{8.1070} & \textbf{7.0885} & \textbf{8.7451} & \textbf{8.4076} & \textbf{6.8414} & \textbf{7.8379} & \textbf{3.0} \\
llama 3\_3 70B\_hier & \textbf{0.3746} & \textbf{0.2327} & \textbf{0.1931} & 0.7730 & 0.4987 & \textbf{0.2863} & 7.3417 & 6.3637 & 8.1545 & 7.6200 & 5.9002 & 7.0760 & 3.714 \\
eurollm 9B\_hier & 0.2800 & 0.1856 & 0.1451 & 0.7459 & 0.5065 & 0.1875 & 4.9739 & 4.4255 & 6.4043 & 6.6876 & 3.5435 & 5.2070 & 8.286 \\
llama 3 8B\_hier & 0.2927 & 0.1829 & 0.1472 & 0.7373 & 0.5082 & 0.2289 & 5.2780 & 4.5405 & 6.3069 & 6.4295 & 3.7751 & 5.2660  & 7.500 \\
mistral v03\_hier & 0.3571 & 0.2304 & 0.1871 & \textbf{0.7777} & \textbf{0.5122} & 0.2386 & 5.5376 & 4.9653 & 5.5578 & 5.2447 & 3.7370 & 5.0085 & 6.929 \\
teuken\_hier\_v0.4 & 0.1630 & 0.0794 & 0.0781 & 0.6600 & 0.5051 & 0.1607 & 3.0635 & 2.1606 & 4.2356 & 4.4077 & 1.8269 & 3.1388 & 10.429\\
teuken\_inst\_hier\_v0.4\_sft & 0.2865 & 0.2181 & 0.1758 & 0.8 & 0.4893 & 0.12 & 3.0758 & 2.0058 & 4.1411 & 4.9223 & 1.5835 & 3.1457 &  10.214 \\ \midrule
teuken\_hier\_v0.6 & 0.2621 & 0.1693 & 0.1325 & 0.7200 & 0.5111 & 0.1644 & 4.5371 & 3.9509 & 4.6766 & 4.8710 & 2.9182 & 4.1908 & n/a \\
teuken\_inst\_hier\_v0.6\_sft & 0.2786 &	0.1837 &	0.1445&	0.7399&	0.5024&	0.1858&	4.6306&	3.9302&	4.5730&	4.9513	&2.8280&	4.2049&n/a \\
\midrule
gpt 4o\_full & 0.3627 & 0.2105 & 0.1845 & 0.7563 & 0.4991 & 0.2777 & \textbf{8.3933} & \textbf{7.1615} & \textbf{8.8192} & \textbf{8.5385} & \textbf{7.0115} & \textbf{7.9848} & \textbf{2.714} \\
llama 3\_3 70B\_full & \textbf{0.3823} & \textbf{0.2248} & \textbf{0.1986} & \textbf{0.7691} & \textbf{0.5082} & 0.2898 & 8.1721 & 6.8661 & 8.6333 & 8.1552 & 6.6603 & 7.6974 & 2.929 \\
mistral v03\_full & 0.3612 & 0.2126 & 0.1901 & 0.7465 & 0.5021 & \textbf{0.3252} & 6.9612 & 5.7141 & 7.1395 & 6.8110 & 5.0271 & 6.3306 & 5.429 \\
\bottomrule
\end{tabular*}
\caption{Quantitative results on hierarchical and full judgments. Human Avg = mean rank across 14 cases; Correlation between LJ\_Tot and Human Average: Spearman's $\rho$ = -0.939 (p < 0.001).}
\label{tab:comparison_merged}
\end{table*}

\section{Evaluation}\label{evaluation}

We developed a comprehensive evaluation approach using
multiple complementary metrics: ROUGE (\citet{lin2004}), BLEU (\citet{papineni2002}), METEOR (\citet{banerjee2005}), BERTScore (\citet{zhang2020}), QAGS
  (\citet{wang2020}), FactCC (\citet{Kryscinski2019}) and LLM-as-judge.

\subsubsection{Factual Consistency
Metrics}\label{factual-consistency-metrics}

Because press releases can include context not explicitly stated in the court decisions, QAGS and FactCC metrics may flag such information as inconsistent, potentially lowering scores for otherwise high-quality press releases. We partially address this through our LLM-as-judge approach and the human evaluation process, which better distinguishes contradictory information from benign additional context.

\subsubsection{LLM-as-judge}\label{LLM-as-judge}

We use Claude 3.7 Sonnet to evaluate the generated press releases based
on completeness, clarity, structure and comparison to the reference. 
The metric provides numerical ratings (1-10) and calculate an overall score
across all evaluation criteria. The model was selected for this task due to its strong performance in understanding
complex legal texts in multiple languages as well as its selection for
synthetic prompt generation which made it a natural choice for evaluation.

\subsection{Human Evaluation}\label{human-evaluation}

In addition to automatic metrics, we conducted a limited human evaluation with three annotators (including a licensed German attorney) who blindly ranked press releases from 1 (best) to 11 (worst) for 2-3 cases per court per model. While this sample size precludes statistically significant conclusions, it provides qualitative directional insights into model performance.

The "Human Avg" score in \hyperref[tab:comparison_merged]{Table 2} represents the mean rank assigned to each system across all annotated cases, with lower values indicating better performance. For example, a Human Avg of 3.0 means that system averaged 3rd place across evaluations.

Annotators also assessed narrative coherence and usability (i.e. whether it is close to publishable) of the generated press release, and whether it contains extraneous information not in the judgment. The latter not only covers hallucinations of LLMs, but also factually correct enriched information. Note that even the reference press release regularly contains dates, case numbers and the social context of the case that might not be represented in the document itself.

\section{Results}\label{results}
Table 2 presents results organized by evaluation type (hierarchical vs. full document) and model, reference-based, embedding-based, and factual consistency metrics, plus LLM-as-judge scores and human rankings.

The full-text condition reveals the upper bound when context is not truncated, while hierarchical setting approximates local-deployment scenarios. GPT-4o and Llama-3-70B achieve comparable automatic metrics, though LLM judging clearly prefers GPT-4o. For Mistral\_v03, we evaluated on full ruling text despite its 32k token limit; only 1\% of documents required truncation, representing negligible noise.

To validate the LLM-as-judge approach, we computed correlations between LLM-judge scores (LJ\_Tot, 1-10 scale, higher=better) and human rankings (1-11, lower=better) for 10 model configurations. Analysis revealed strong alignment: Spearman's $\rho$ = -0.939, Kendall's $\tau$ = -0.822, and Pearson's r = -0.987 (all p < 0.001). These significant negative correlations where higher LLM scores correspond to better (lower) human rankings provide evidence of LLM-expert alignment, despite limited human evaluation. The ordering of the overall automatic quality score corresponds to the average human ranking with the exception of the Mistral-7B model, which ranks third in human scores but only fifth in automatic scores.

\section{Discussion}\label{discussion}
Our results align with \citet{glaser2021}, who reported ROUGE-1 scores around 30.5\% for German court decision summarization due to LLM advances.

The findings confirm expected trade-offs: large models (GPT-4o, Llama-3-70B) substantially outperform smaller ones on fidelity, completeness, and clarity, though this gap narrows with hierarchical summarization. LLM-as-judge shows encouraging alignment with expert feedback. Annotators consistently ranked reference press releases highest, while larger models' outputs appeared to need only minor edits for practical use.

Hierarchical summarization enables smaller models to produce reasonably good summaries, particularly for clarity and structure. The improvement from hierarchical to full summarization is marginal for largest models. Our fine-tuned Teuken model, despite improvement, performs far below larger models, confirming parameter count remains decisive.

The factual consistency scores (QAGS, FactCC) require cautious interpretation given their English-language origins. Low scores may reflect limited transferability to German legal language rather than factual errors, underscoring the value of our LLM-as-judge approach for German-specific terminology.

\section{Conclusions}
CourtPressGER demonstrates that modern LLMs can effectively generate German court press releases, with performance metrics strongly correlating with model size and architecture. Larger models consistently outperform smaller ones across all metrics, while hierarchical summarization successfully enables smaller models to process long documents with reasonable quality preservation. The most significant challenge remains factual consistency, where even top-performing models struggle, suggesting this as a critical area for future research. Notably, European-specific models, like EuroLLM, demonstrate competitive performance relative to their size, indicating the value of jurisdiction-aware training. These findings contribute to further work in developing and evaluating systems targeted at communicating court judgment to a wider audience while highlighting the continued superiority of human-drafted press releases, which consistently ranked highest in expert evaluation.

\section{Limitations}\label{limitations}

We found several limitations of our approach:

\begin{enumerate}
\def\labelenumi{\arabic{enumi}.}
\item
  LLM-as-judge vs.~human evaluation: While our LLM-based evaluation
  provides valuable insights, it serves as a proxy for human expert
  evaluation and would benefit from further validation through targeted expert
  reviews.
\item
  Additional context in press releases: Court press releases often
  contain contextual information not present in the original decision,
  which can confound factual consistency metrics.
\item
  Divergence from Rolshoven et al.~findings: Unlike \citet{rolshoven2024}, who found that fine-tuned smaller models could approach
  the performance of larger models, our results show a clear advantage
  for larger models. This difference may be attributed to our focus on
  press releases rather than technical summaries (``Regesten''), the
  different nature of our dataset, or the specific characteristics of
  German federal court decisions.
\item
  Limited human evaluation scope: While our correlation analysis shows very strong agreement ($\rho$ = -0.939) between LLM-judge and human rankings, this is based on only 10 model configurations and 14 human-evaluated cases total. Larger-scale human evaluation would provide more robust validation of these promising initial findings.
\item
  Evaluation metrics: Our use of QAGS and FactCC metrics, which were developed and validated on English datasets, introduces uncertainty when applied to German legal texts. These metrics may not adequately handle German compound words, case-based grammar, or jurisdiction specific legal terminology. Additionally, the language gap could lead to systematic biases in factual consistency scoring. Future work should prioritize developing German-specific factual consistency metrics, potentially leveraging German question answering models and legal domain knowledge.
\item 
  Synthetic prompt validation: While our use of Claude-generated prompts may introduce bias, we maintain consistent prompting across all models to ensure fair relative comparison. Future work should investigate the impact of prompt generation methods through ablation studies comparing synthetic versus generic prompts.
  
\end{enumerate}

\section{Ethics Statement}\label{ethics-statement}

All data originate from publicly available court websites. Personal
names are already anonymised by the courts. Our dataset is released excluding any confidential
meta‑data.

\section{AI usage}\label{ai-usage}

We leveraged Claude Sonnet 3.7 for coding tasks and GPT-4o for wording, shortening and Latex tasks.

\section{Acknowledgements}
This project has been funded partially by a research collaboration with the SINC GmbH and by the German Federal Ministry of Justice as part of the project ``Generatives Sprachmodell der Justiz''. We thank Klaas Schmidt, Angelina Greiner, and Rusheel Iyer for their valuable contributions and support in the development of this work.

\clearpage

\bibliography{CourtPressGER}

\appendix\label{appendix}
\section*{Appendix}
\section{Prompts - Original}\label{prompts}

We used the following prompts for our experiments:

\subsection{Synthetic prompt
generation}\label{synthetic-prompt-generation}

We used the following prompt for synthetic prompt generation: 

\begin{itshape}
Du bist ein Experte für juristische Texte und Kommunikation. Deine
Aufgabe ist es, ein Gerichtsurteil und die dazugehörige Pressemitteilung
zu analysieren und dann herauszufinden, welcher Prompt verwendet worden
sein könnte, um diese Pressemitteilung aus dem Gerichtsurteil zu
generieren, wenn man ihn einem LLM gegeben hätte.

\begin{enumerate}
\def\labelenumi{\arabic{enumi}.}
\item
  Analysiere, wie die Pressemitteilung Informationen aus dem Urteil
  vereinfacht, umstrukturiert und Schlüsselinformationen hervorhebt
\item
  Berücksichtige den Ton, die Struktur und den Detaillierungsgrad der
  Pressemitteilung
\item
  Identifiziere, welche Anweisungen nötig wären, um den juristischen
  Text in diese Pressemitteilung zu transformieren
\end{enumerate}

Erkläre NICHT deine Überlegungen und füge KEINE Meta-Kommentare hinzu.
Gib NUR den tatsächlichen Prompt aus, der die Pressemitteilung aus dem
Gerichtsurteil generieren würde. Sei spezifisch und detailliert in
deinem synthetisierten Prompt.

Hier ist das originale Gerichtsurteil: \{court\_ruling\}

Und hier ist die Pressemitteilung, die daraus erstellt wurde:

\{press\_release\}

Erstelle einen detaillierten Prompt, der einem LLM gegeben werden
könnte, um die obige Pressemitteilung aus dem Gerichtsurteil zu
generieren. Schreibe NUR den Prompt selbst, ohne Erklärungen oder
Meta-Kommentare.
\end{itshape}

\subsubsection{Press release
generation}\label{press-release-generation-1}

We used the following prompt for press release generation:

\begin{itshape}

\{prompt\} 

Gerichtsurteil: 

\{ruling\}

\end{itshape}

\subsection{LLM-as-judge}\label{LLM-as-judge-1}

We used the following prompt for LLM-as-judge evaluation:

\begin{itshape}
        Du bist ein Experte für juristische Texte und bewertst die Qualität von Pressemitteilungen für Gerichtsurteile.
        Bewerte die generierte Pressemitteilung anhand der folgenden Kriterien auf einer Skala von 1-10:
        
        1. Faktische Korrektheit: Wie genau spiegelt die Pressemitteilung die Fakten aus dem Gerichtsurteil wider?
        
        2. Vollständigkeit: Wurden alle wichtigen Informationen aus dem Urteil in der Pressemitteilung berücksichtigt?
        
        3. Klarheit: Wie verständlich ist die Pressemitteilung für ein nicht-juristisches Publikum?
        
        4. Struktur: Wie gut ist die Pressemitteilung strukturiert und organisiert?
        
        5. Vergleich mit Referenz: Wie gut ist die generierte Pressemitteilung im Vergleich zur Referenz-Pressemitteilung?
        
        Gib für jedes Kriterium einen numerischen Wert zwischen 1 und 10 an und eine kurze Begründung.
        Berechne abschließend einen Gesamtscore als Durchschnitt aller Einzelwerte.
        Gib deine Antwort im folgenden JSON-Format zurück:
        
        \{
        
            "faktische\_korrektheit": \{
            
            "wert": X, "begründung": "..."
            
            \},

            "vollständigkeit": \{
            
            "wert": X, "begründung": "..."
            
            \},
            "klarheit": \{
            
            "wert": X, "begründung": "..."
            
            \},
            "struktur": \{
            
            "wert": X, "begründung": "..."
            
            \},
            
            "vergleich\_mit\_referenz": \{
            
            "wert": X, "begründung": "..."
            
            \},
            
            "gesamtscore": X.X
            
        \}

        \# Gerichtsurteil
        
        \{source\}
        
        \# Generierte Pressemitteilung
        
        \{generated\}
        
        \# Referenz-Pressemitteilung
        
        \{reference\}
\end{itshape}

\section{Prompts - English Translations}\label{english-translations}

For international readers, we provide English translations of the prompts used in our experiments:

\subsection{Synthetic prompt
generation - Translated}\label{synthetic-prompt-generation}

\begin{itshape}

You are an expert in legal texts and communication. Your task is to analyze a court ruling and its corresponding press release, and then determine what prompt could have been used to generate this press release from the court ruling if it had been given to an LLM.

\begin{enumerate}
\def\labelenumi{\arabic{enumi}.}
\item
  Analyze how the press release simplifies, restructures, and highlights key information from the ruling
\item
  Consider the tone, structure, and level of detail of the press release
\item
  Identify what instructions would be necessary to transform the legal text into this press release
\end{enumerate}

Do NOT explain your reasoning and do NOT add meta-comments. Output ONLY the actual prompt that would generate the press release from the court ruling. Be specific and detailed in your synthesized prompt.

Here is the original court ruling: \{court\_ruling\}

And here is the press release that was created from it:

\{press\_release\}

Create a detailed prompt that could be given to an LLM to generate the above press release from the court ruling. Write ONLY the prompt itself, without explanations or meta-comments.
\end{itshape}

\subsection{Press release generation - Translated}\label{synthetic-prompt-generation}

\begin{itshape}

\{prompt\} 

Court ruling: 

\{ruling\}

\end{itshape}

\subsection{LLM-as-judge - Translated}\label{synthetic-prompt-generation}

\begin{itshape}

You are an expert in legal texts and evaluate the quality of press releases for court rulings.
Evaluate the generated press release based on the following criteria on a scale of 1-10:

1. Factual Correctness: How accurately does the press release reflect the facts from the court ruling?

2. Completeness: Were all important information from the ruling considered in the press release?

3. Clarity: How understandable is the press release for a non-legal audience?

4. Structure: How well structured and organized is the press release?

5. Comparison with Reference: How good is the generated press release compared to the reference press release?

For each criterion, provide a numerical value between 1 and 10 and a brief justification.
Finally, calculate an overall score as the average of all individual values.
Return your answer in the following JSON format:

\{

    "factual\_correctness": \{
    
        "value": X, "justification": "..."
        
    \},
    
    "completeness": \{
    
        "value": X, "justification": "..."
        
    \},
    
    "clarity": \{
    
        "value": X, "justification": "..."
        
    \},
    
    "structure": \{
    
        "value": X, "justification": "..."
        
    \},
    
    "comparison\_with\_reference": \{
    
        "value": X, "justification": "..."
        
    \},
    
    "overall\_score": X.X
    
\}

\# Court ruling

\{source\}

\# Generated press release

\{generated\}

\# Reference press release

\{reference\}

\end{itshape}

\section{Full evaluation results}\label{full evaluataion results}

\noindent\footnotesize
\begin{tabular}{@{}ll@{\hspace{3em}}ll@{}}
R1, R2, RL   & ROUGE-1/-2/-L F1            & KW          & Keyword-Overlap\\
B1–B4        & BLEU-1 … BLEU-4             & ENT         & Entity-Overlap\\
MTR          & METEOR                      & Len         & Length-Ratio\\
BP, BR, BF1  & BERTScore Precision/Recall/F1 & Fcc, FccC   & FactCC Score / Consistency\\
QGS, Qn      & QAGS Score / Ø Questions       & LJ\_Fact    & \texttt{llm\_judge}\,fact. Corr.\\
             &                             & LJ\_Compl   & LLM-as-judge Completeness\\
             &                             & LJ\_Clar    & LLM-as-judge Clarity\\
             &                             & LJ\_Struc   & LLM-as-judge Structure\\
             &                             & LJ\_Ref     & LLM-as-judge Comparison with Reference\\
             &                             & LJ\_Tot     & LLM-as-judge Total Score\\
\end{tabular}

\begin{table}[!htpb]
\scriptsize
\setlength{\tabcolsep}{6pt}
\renewcommand{\arraystretch}{1.05}
\centering
\begin{tabular}{lrrrrrrrrrrrr}
\toprule
Model &
R1 & R2 & RL &
B1 & B2 & B3 & B4 &
MTR &
BP & BR & BF1
\\ \midrule
openai\_gpt\_4o\_full &
0.3627 & 0.1452 & 0.1918 &
0.2105 & 0.1266 & 0.0832 & 0.0559 &
0.1845 &
0.7746 & 0.7396 & 0.7563
\\
openai\_gpt\_4o\_hier &
0.3584 & 0.1242 & 0.1758 &
0.2275 & 0.1280 & 0.0786 & 0.0495 &
0.1836 &
0.7835 & 0.7595 & 0.7711
\\
llama\_3\_3\_70B\_full &
\textbf{0.3823} & \textbf{0.1601} & \textbf{0.1997} &
0.2248 & \textbf{0.1385} & \textbf{0.0946} & \textbf{0.0668} &
\textbf{0.1986} &
0.7889 & 0.7508 & 0.7691
\\
llama\_3\_3\_70B\_hier &
0.3746 & 0.1411 & 0.1864 &
\textbf{0.2327} & 0.1358 & 0.0879 & 0.0593 &
0.1931 &
\textbf{0.7918} & 0.7557 & 0.7730
\\
eurollm\_9B\_hier &
0.2800 & 0.0611 & 0.1199 &
0.1856 & 0.0832 & 0.0413 & 0.0212 &
0.1451 &
0.7570 & 0.7362 & 0.7459
\\
llama\_3\_8B\_hier &
0.2927 & 0.0780 & 0.1344 &
0.1829 & 0.0897 & 0.0499 & 0.0287 &
0.1472 &
0.7519 & 0.7239 & 0.7373
\\
mistral\_v03\_full &
0.3612 & 0.1561 & 0.1844 &
0.2126 & 0.1304 & 0.0907 & 0.0660 &
0.1901 &
0.7706 & 0.7255 & 0.7465
\\
mistral\_v03\_hier &
0.3571 & 0.1218 & 0.1638 &
0.2304 & 0.1264 & 0.0780 & 0.0509 &
0.1871 &
\textbf{0.7918} & \textbf{0.7645} & \textbf{0.7777}
\\
teuken\_hier &
0.1630 & 0.0213 & 0.0703 &
0.0794 & 0.0284 & 0.0105 & 0.0043 &
0.0781 &
0.6966 & 0.6303 & 0.6600
\\
teuken\_inst\_sft &
0.2865 & 0.0569 & 0.1164 &
0.2181 & 0.0993 & 0.0531 & 0.0312 &
0.1758 &
0.7973 & 0.8037 & 0.8000
\\
\bottomrule
\end{tabular}
\caption{Automatic reference-based metrics (ROUGE, BLEU, METEOR, BERTScore).}
\label{tab:auto_a}
\end{table}
\begin{table}[!htpb]
\scriptsize
\setlength{\tabcolsep}{15pt}
\renewcommand{\arraystretch}{1.05}
\centering
\begin{tabular}{lrrrrrrrr}
\toprule
Model &
KW & ENT & Len &
Fcc & FccC &
QGS & Qn
\\ \midrule
openai\_gpt\_4o\_full &
0.2082 & 0.2290 & 0.4572 &
0.4991 & 0.5068 &
0.2777 & 4.75
\\
openai\_gpt\_4o\_hier &
0.1883 & 0.2157 & 0.5114 &
0.4915 & 0.4758 &
0.2637 & 4.78
\\
llama\_3\_3\_70B\_full &
\textbf{0.2198} & \textbf{0.2311} & 0.4972 &
0.5082 & 0.5144 &
0.2898 & 4.87
\\
llama\_3\_3\_70B\_hier &
0.2132 & 0.2158 & 0.5156 &
0.4987 & 0.5005 &
0.2863 & \textbf{4.94}
\\
eurollm\_9B\_hier &
0.1275 & 0.1229 & 0.5249 &
0.5065 & \textbf{0.5290} &
0.1875 & 4.84
\\
llama\_3\_8B\_hier &
0.1456 & 0.1444 & 0.4958 &
0.5082 & 0.5081 &
0.2289 & 4.90
\\
mistral\_v03\_full &
0.2132 & 0.2074 & 0.4929 &
0.5021 & 0.5044 &
\textbf{0.3252} & 4.72
\\
mistral\_v03\_hier &
0.1884 & 0.1825 & \textbf{0.5475} &
\textbf{0.5122} & 0.5189 &
0.2386 & 4.69
\\
teuken\_hier &
0.0705 & 0.0673 & 0.3553 &
0.5051 & 0.5068 &
0.1607 & \textbf{4.94}
\\
teuken\_inst\_sft &
0.1332 & 0.1317 & 0.6039 &
0.4893 & 0.4542 &
0.1200 & 4.54
\\
\bottomrule
\end{tabular}
\caption{Automatic content-overlap, factuality, and answerability metrics.}
\label{tab:auto_b}
\end{table}


\vspace{1em}

\begin{table}[!htpb]
\scriptsize
\setlength{\tabcolsep}{14pt}
\renewcommand{\arraystretch}{1.05}
\centering
\begin{tabular}{lrrrrrr}
\toprule
Model &
LJ\_Fact & LJ\_Compl & LJ\_Clar & LJ\_Struc & LJ\_Ref & LJ\_Tot
\\ \midrule
openai\_gpt\_4o\_full &
\textbf{8.3933} & \textbf{7.1615} & \textbf{8.8192} & \textbf{8.5385} & \textbf{7.0115} & \textbf{7.9848}
\\
openai\_gpt\_4o\_hier &
8.1070 & 7.0885 & 8.7451 & 8.4076 & 6.8414 & 7.8379
\\
llama\_3\_3\_70B\_full &
8.1721 & 6.8661 & 8.6333 & 8.1552 & 6.6603 & 7.6974
\\
llama\_3\_3\_70B\_hier &
7.3417 & 6.3637 & 8.1545 & 7.6200 & 5.9002 & 7.0760
\\
eurollm\_9B\_hier &
4.9739 & 4.4255 & 6.4043 & 6.6876 & 3.5435 & 5.2070
\\
llama\_3\_8B\_hier &
5.2780 & 4.5405 & 6.3069 & 6.4295 & 3.7751 & 5.2660
\\
mistral\_v03\_full &
6.9612 & 5.7141 & 7.1395 & 6.8110 & 5.0271 & 6.3306
\\
mistral\_v03\_hier &
5.5376 & 4.9653 & 5.5578 & 5.2447 & 3.7370 & 5.0085
\\
teuken\_hier &
3.0635 & 2.1606 & 4.2356 & 4.4077 & 1.8269 & 3.1388
\\
teuken\_inst\_sft &
3.0758 & 2.0058 & 4.1411 & 4.9223 & 1.5835 & 3.1457
\\ \bottomrule
\end{tabular}
\caption{LLM-as-judge evaluation scores
         (fact, completeness, clarity, structure, reference comparison, and total).}
\label{tab:llm_judge}
\end{table}

\noindent\footnotesize
\begin{tabular}{@{}ll@{}}
LJ\_Fact  & \texttt{llm\_judge} factual correctness\\
LJ\_Compl & LLM-as-judge completeness\\
LJ\_Clar  & LLM-as-judge clarity\\
LJ\_Struc & LLM-as-judge structure\\
LJ\_Ref   & LLM-as-judge comparison with reference\\
LJ\_Tot   & LLM-as-judge total score\\
\end{tabular}

\vspace{1em}

\begin{table*}[!htpb]
\centering
\begin{tabular}{lrrrr}
\toprule
\textbf{Model}  & \textbf{Avg.\ Rank} & \textbf{Ext.\ Info Rate} & \textbf{Incoherent Rate} & \textbf{Publishable Rate} \\
\midrule
reference\_summary                                          & 1.500  & 0.786 & 0.071 & 1.000 \\
openai\_gpt\_4o\_generated\_full                           & 4.000  & 0.643 & 0.000 & 0.714 \\
llama\_3\_3\_70B\_generated\_full                          & 4.071  & 0.714 & 0.071 & 0.571 \\
openai\_gpt\_4o\_gen\_hier                                 & 4.500  & 0.571 & 0.071 & 0.786 \\
llama\_3\_3\_70B\_gen\_hier                                & 4.714  & 0.714 & 0.000 & 0.571 \\
mistral\_v03\_generated                                    & 5.857  & 0.714 & 0.214 & 0.214 \\
llama\_3\_8b\_gen\_hier                                    & 6.429  & 0.714 & 0.071 & 0.214 \\
eurollm\_gen\_hier                                         & 7.143  & 0.786 & 0.286 & 0.214 \\
mistral\_v03\_gen\_hier                                    & 7.714  & 1.000 & 0.143 & 0.143 \\
teuken\_gen\_hier\_sft                                     & 9.857  & 0.929 & 0.214 & 0.071 \\
teuken\_gen\_hier                                          & 10.214 & 0.714 & 0.214 & 0.000 \\
\bottomrule
\end{tabular}
\caption{Human evaluation of summary quality for each model (hierarchical summaries $\to$ \texttt{\_hier};
          complete judgements $\to$ \texttt{\_full}).}
\label{tab:model_summary}
\end{table*}
\end{document}